\documentclass[11pt]{amsart}
\usepackage{amsfonts}
\usepackage{amsmath}
\usepackage{amssymb}
\usepackage{amsxtra}
\usepackage{amsthm}                
\usepackage[ruled,vlined]{algorithm2e}

\usepackage{comment}
\usepackage{enumerate}
\usepackage{epsfig}
\usepackage{epstopdf}
\usepackage[mathscr]{euscript}
\usepackage{float}
\usepackage[margin=1in]{geometry}  
\usepackage{graphics}
\usepackage{graphicx}
\usepackage{hyperref}
\usepackage[utf8]{inputenc}
\usepackage{latexsym}
\usepackage{mathdots}
\usepackage{microtype}
\usepackage{pict2e}
\usepackage{setspace}
\usepackage{subfigure}
\usepackage{tikz}
\usepackage{wrapfig}

\usepackage{hyperref}

\theoremstyle{definition}
\newtheorem{theorem}{Theorem}[section]
\newtheorem{lemma}[theorem]{Lemma}

\theoremstyle{definition}

\theoremstyle{remark}

\numberwithin{equation}{section}

\usepackage{cite}
\numberwithin{equation}{section}

\newcommand{\Net}{\mathrm{Net}}
\newcommand{\R}{\mathbb{R}}
\newcommand{\din}{{d_{\text{in}}}}
\newcommand{\dout}{{d_{\text{out}}}}

\usepackage{adjustbox}

\begin{document}

\title{Algebraically-Informed Deep Networks (AIDN): \\ A Deep Learning Approach to Represent Algebraic Structures}

\author{Mustafa Hajij}
\address{Department of Mathematics and Computer Science,
Santa Clara University,
Santa Clara, CA USA}
\email{mhajij@scu.edu}

\author{Ghada Zamzmi}
\address{University of South Florida}
\email{ghadh@mail.usf.edu}

\author{Matthew Dawson}
\address{CONACYT Research Fellow, Centro de Investigación en Matemáticas, Mérida Campus, Mérida, Yucatán, México}
\email{matthew.dawson@cimat.mx}

\author{Greg Muller}
\address{Department of Mathematics,
University of Oklahoma,
Norman, OK USA}
\email{gmuller@ou.edu}




 \maketitle


\begin{abstract}
 One of the central problems in the interface of deep learning and mathematics is that of building learning systems that can automatically uncover underlying mathematical laws from observed data. In this work, we make one step towards building a bridge between algebraic structures and deep learning, and introduce \textbf{AIDN}, \textit{Algebraically-Informed Deep Networks}. \textbf{AIDN} is a deep learning algorithm to represent any finitely-presented algebraic object with a set of deep neural networks. The deep networks obtained via \textbf{AIDN} are \textit{algebraically-informed} in the sense that they satisfy the algebraic relations of the presentation of the algebraic structure that serves as the input to the algorithm. Our proposed network can robustly compute linear and non-linear representations of most finitely-presented algebraic structures such as groups, associative algebras, and Lie algebras. We evaluate our proposed approach and demonstrate its applicability to algebraic and geometric objects that are significant in low-dimensional topology. In particular, we study solutions for the Yang-Baxter equations and their applications on braid groups. Further, we study the representations of the Temperley-Lieb algebra. Finally, we show, using the Reshetikhin-Turaev construction, how our proposed deep learning approach can be utilized to construct new link invariants. We believe the proposed approach would tread a path toward a promising future research in deep learning applied to algebraic and geometric structures. 
  
 
\end{abstract}

\section{Introduction}

Over the past years, deep learning techniques have been used for solving partial differential equations
\cite{lagaris1998artificial,lagaris2000neural,raissi2018deep,sirignano2018dgm}, obtaining physics-informed surrogate models \cite{raissi2017physics,rudy2017data}, computing the Fourier transform with deep networks \cite{li2020fourier}, finding roots of polynomials \cite{huang2004neural}, and solving non-linear implicit system of equations \cite{song2020nonlinear}. In this work, we make one step towards building the bridge between algebraic/geometric structures and deep learning, and aim to answer the following question: How can deep learning be used to uncover the underlying solutions of an arbitrary system of algebraic equations?

To answer this question, we introduce \textit{Algebraically-Informed Deep Networks} (\textbf{\textbf{AIDN}}), a deep learning method to represent any finitely-presented algebraic object with a set of neural networks. Our method uses a set of deep neural networks to represent a set of formal algebraic symbols that satisfy a system of algebraic relations. These deep neural networks are simultaneously trained to satisfy the relations between these symbols using an optimization paradigm such as stochastic gradient descent (SGD). The resulting neural networks are \textit{algebraically-informed} in the sense that they satisfy the algebraic relations. We show that a wide variety of mathematical problems can be solved using this formulation. 

Next, we discuss a motivating example and present the applicability of our method on the well-known Yang-Baxter equation \cite{jimbo1989introduction}, which has been extensively studied in mathematics and physics.

\subsection{Motivating Example: The Yang-Baxter Equation}
\label{ybe}

To solve the set-theoretic Yang-Baxter equation, one seeks an invertible function $R:A\times A\ \to A\times A $, where $A$ is some set, that satisfies the following equation: 
\begin{equation}
\label{yb_eq}
 (R \times id_{A})\circ (id_{A} \times R)\circ (R \times id_{A}) = (id_{A} \times R) \circ (R \times id_{A})\circ  (id_{A} \times R).
\end{equation}
Finding solutions of the above equation has a long history and proved to be a very difficult problem. For many decades, the Yang–Baxter equation\footnote{Technically, the term \textit{Yang-Baxter equation} is utilized whenever the map $R$ is linear. When the map $R$ is an arbitrary map defined on a set, the term \textit{set-theoretic Yang-Baxter} is used instead.} has been studied in quantum field theory and statistical mechanics as the master equation in integrable models \cite{jimbo1989introduction}. Later, this equation was applied to many problems in low-dimensional topology \cite{kassel2008braid}. For example, solutions of the Yang-Baxter equation were found to induce representations of the braid groups and have been used to define knots and 3-manifold invariants \cite{turaev2020quantum}. Today, the Yang-Baxter equation is considered a cornerstone in several areas of physics and mathematics \cite{vieira2018solving} with applications to quantum mechanics \cite{kauffman2004braiding}, algebraic geometry \cite{krichever1981baxter}, and quantum groups \cite{turaev1988yang}.

As an example application, we show how the proposed \textbf{\textbf{AIDN}} can be utilized to solve the Yang-Baxter equation \cite{etingof1998set}. In particular, assuming $A\subseteq\R^n$ is a subset of a Euclidean space\footnote{We may consider real or complex Euclidean spaces, but for this example we will constrain our discussion on real-Euclidean spaces.}, \textbf{\textbf{AIDN}} realizes the desired solution $R$ in equation \ref{yb_eq} as a neural network $f_R(\theta)$, where $\theta \in \R^k$. Using SGD, we can find the parameters $\theta$ by optimizing a loss function, which essentially satisfies equation \ref{yb_eq}. More details are given in Section \ref{braidgroupsection}.

\subsection{Related Work}
Our work can be viewed as a part of the quest to discover knowledge and a step towards building learning systems that are capable of uncovering the underlying mathematical and physical laws from data. Examples of current deep learning efforts to solve problems in mathematics and physics include general methods to solve partial differential equations \cite{sirignano2018dgm,lagaris1998artificial,lagaris2000neural,raissi2018deep} or more particular ones that are aimed at solving single equations such as the Schrödinger equation \cite{mills2017deep}. Also, deep learning has been used to solve equations related to fluid mechanics \cite{brunton2020machine,wang2020towards}, non-linear equations \cite{mathia1995solving,song2020nonlinear} and transcendental equations \cite{jeswal2018solving}.

This work can also be viewed as a step towards advancing computational algebra \cite{seress1997introduction,lux2010representations}. Although there is a large literature devoted to computing linear representations of finitely-presented algebraic objects \cite{holt2005handbook} and of finite groups in particular \cite{steel2012construction,dabbaghian2005algorithm} as well as few works about representation of algebras \cite{fischbacher1986algorithms}, we are not aware of any algorithm that computes non-linear representations of algebraic structures. Further, existing works find the representations of algebraic structures in special cases \cite{adams2008algorithms}; the majority of these algorithms utilize GAP \cite{gap2007gap}, a system for computational discrete algebra. Our proposed \textbf{\textbf{AIDN}} can (1) compute both linear and non-linear representations of algebraic structures, (2) provide a general computational scheme that utilizes non-traditional tools, and (3) offer a different paradigm from the classical methods in this space.

\subsection{Summary of Contribution}
The main contributions of this work can be summarized as follows:
\begin{enumerate}
    \item We propose \textbf{AIDN}, a deep learning algorithm that computes non-linear representations of algebraic structures. To the best of our knowledge, we are the first to propose a deep learning-based method for computing non-linear representations of any finitely presented algebraic structure.

    \item We demonstrate the applicability of \textbf{AIDN} in low-dimensional topology. Specifically, we study the applicability of \textbf{AIDN} to braid groups and Templery-Lieb algebras, two algebraic constructions that are significant in low-dimensional topology.
    
    \item We utilize \textbf{\textbf{AIDN}} for knot invariants discovery using deep learning methods. Specifically, using the Reshetikhin-Turaev construction we show that \textbf{AIDN} can be used to construct new link invariants.  
\end{enumerate}

The rest of the paper is organized as follows. Section \ref{nets} reviews the basics of neural networks that are needed for our settings. In Section \ref{method}, we present our \textbf{\textbf{AIDN}} method. In Section \ref{cases}, we study the application of \textbf{\textbf{AIDN}} algorithm to the braid groups and Templery-Lieb algebras. In Section \ref{knots} we show how \textbf{AIDN} can be utilized to obtain invariants of knots and links. Finally, we discuss the limitations and conclude the paper in Section \ref{final}. 









\section{Background: Algebraic Structures with Deep Neural Networks}
\label{nets}


This section provides a brief introduction to neural networks and shows examples of the algebraic structures that can be defined on them. We only focus on real-neural networks with domains and co-domains on real Euclidean spaces for the sake of clarity and brevity. However, \textbf{\textbf{AIDN}} can be easily extended to complex-neural networks \cite{hirose2003complex,kim2002universal}.

A \textit{neural network}, or simply \textit{network}, is a function $\Net: \R^\din \longrightarrow \R^\dout$ defined by a composition of the form: 
\begin{equation}
\label{Net}
   \Net:=f_{L} \circ \cdots \circ f_{1}
\end{equation}
where the functions $f_{i}$, $1 \leq i \leq L $ called the \textit{layer functions}. A layer function $f_i:\R^{n_i} \longrightarrow \R^{m_i} $ is typically a continuous, piecewise smooth, or smooth function of the following form: $f_i(x)=\alpha_i (W_i(x)+b_i)$, where $W_i$ is  an $m_i\times n_i$ matrix, $b_i$ is a vector in $\R^{m_i} $, and $\alpha_i :\R \to \R $ is an appropriately chosen nonlinear function applied coordinate-wise to an input vector $(z_1,\cdots, z_{m_i} ) $ to get a vector $(\alpha( z_1),\cdots, \alpha(z_{m_i}))$.

\begin{figure}[t]
  \centering
   {\includegraphics[scale=0.33]{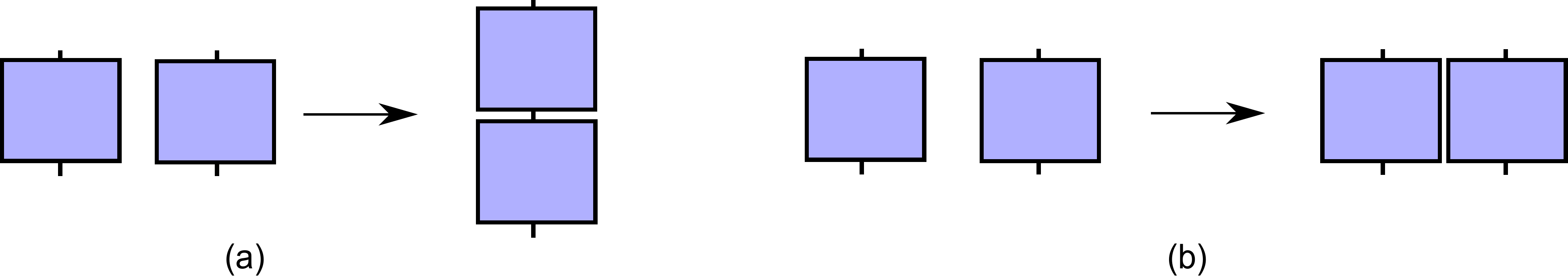}
      \put (-330,35) {\small{$\Net_2$}}
    \put (-368,35) {\small{$\Net_1$}}
     \put (-255,48) {\small{$\Net_1$}}
     \put (-255,21) {\small{$\Net_2$}}
          \put (-342,35) {\small{$\circ$}}
      \put (-135,35) {\small{$\Net_2$}}
    \put (-177,35) {\small{$\Net_1$}}
     \put (-25,35) {\small{$\Net_2$}}
     \put (-55,35) {\small{$\Net_1$}}
          \put (-149,35) {\small{$\times$}}          \caption{ (a) Composition of two nets; (b) product of two nets. }
  \label{product}}
\end{figure}
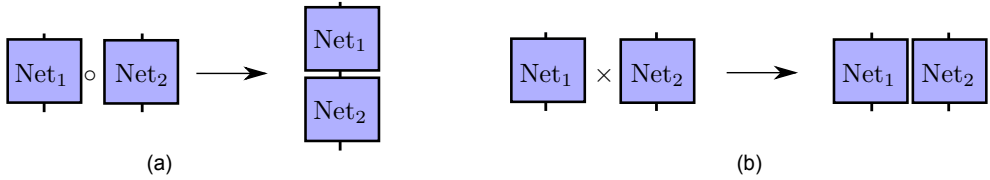

We will use $\mathcal{N}(\R^n)$ to denote the set of networks of the form $\Net : \R^n \to \R^n$. Note that $\mathcal{N}(\R^n)$ is closed under composition of functions. If $ \Net_1 \in  \mathcal{N}(\R^m)$ and $ \Net_2 \in  \mathcal{N}(\R^n)$, then we define $\Net_1 \times \Net_2 \in \mathcal{N}(\R^m \times \R^n  )   $ via $(\Net_1 \times \Net_2 )(x,y) := (\Net_1(x), \Net_2(y)) $ for $x \in \R^n$ and $ y \in \R^m$. We often use the graphical notation illustrated in Figure \ref{product} to denote the composition and the product operations on networks. 

The set $\mathcal{N}(\R^n) $, or a subset of it, admits natural algebraic structures as follows. For two network $\Net_1, \Net_2 \in \mathcal{N}(\R^n)$, we can define their addition $\Net_1+\Net_2$ simply by $(\Net_1+\Net_2)(x)=\Net_1(x)+\Net_2(x)$ for every $x \in \R^n $. Similarly, if $a\in \R$ then  $a*\Net$ is defined via $(a*\Net)(x)=a*\Net(x)$. Function composition, addition, and scalar multiplication defines an associative $\R$-algebra structure on $\mathcal{N}(\R^n) $. We will denote the group inside $\mathcal{N}(\R^n) $ of all invertible networks by $\mathcal{G}(\R^n) $. Finally, we can define, on the associative algebra $\mathcal{N}(\R^n)$, a Lie algebra structure by defining the Lie bracket as $[\Net_1,\Net_2]:=\Net_1 \circ \Net_2- \Net_2 \circ \Net_1$. Given the above setting, we can now represent many types of algebraic structures inside $\mathcal{N}(\R^n) $. 

\section{Method: Algebraically-Informed Deep Networks (\textbf{AIDN})}
\label{method}
The motivation example provided in Section \ref{ybe} can be defined formally and generally as follows. Let $s_1, \ldots, s_n $ be a collection of formal symbols (generators) that satisfy a system of equations $r_1, \ldots, r_k$ which describe a formal set of equations these generators satisfy. We are interested in finding functions $f_{s_1}, \ldots f_{s_n} $ (defined on some domain) that correspond to the formal generators $s_1, \ldots s_n $ and satisfy the same relations $r_1,\ldots, r_k$.

Let the sets $\{s_i\}_{i=1}^n$ and $\{r_i\}_{i=1}^k$ be denoted by $S$ and $R$, respectively. Such a system $
\langle S \mid R\rangle$ is called a \textit{presentation}. Depending on the algebraic operations that we are willing to allow while solving these algebraic equations, presentations can encode different algebraic objects. For example, if we allow multiplication between the algebraic objects and require that this multiplication be associative, have a multiplicative identity, and admit multiplicative inverses for all objects, then the resulting algebraic structure induced by the presentation $\langle S \mid R \rangle$ is a group, and the presentation can be thought of as a compact definition of the group\footnote{One should keep in mind that a group can in general have different presentations.  Furthermore, even the problem of determining which presentations give the trivial group is known to be algorithmically undecidable \cite{lyndon1977combinatorial,rabin1958}.}. On the other hand, finding functions $\{f_{i}\}_{i=1}^n $ that correspond to the generators $S$ and satisfy the relations $R$ is formally equivalent to finding a homomorphism from the algebraic structure $\langle S \mid R \rangle$ to another algebraic structure where the functions $\{f_{i}\}_{i=1}^n$ live.    

From this perspective, \textbf{AIDN} can be formally thought of as taking an algebraic structure given as a finite presentation; that is, a set of generators $\{s_i\}_{i=1}^n$ and a set of relations $\{r_i\}_{i=1}^k$, and producing a set of neural nets $\{f_{i}(x;\theta_{i}) \}_{i=1}^n $, where $\theta_i \in \R^{k_i} $ is the parameter vector of the network $f_i$, such that these neural nets correspond to the generators $\{s_i\}_{i=1}^n$ and satisfy the relations $\{r_i\}_{i=1}^k$. The proposed \textbf{AIDN} finds the weights $\{\theta_{i}\}_{i=1}^n$ of the networks  $\{f_{i}(x;\theta_{i}) \}_{i=1}^n $ by defining the loss function as follows:
 \begin{equation}
     \mathcal{L}(f_1,\cdots,f_n):= \sum_{i=1}^k ||\mathcal{F}(r_i)||_{2}^2,  
 \end{equation}
where $\mathcal{F}(r_i)$ is the relation $r_i$ written in terms of the networks $\{f_{i}(x;\theta_{i}) \}$ and $||.||_2$ is the $L^2$ norm. This loss can be minimized using one of the versions of SGD \cite{bottou2012stochastic}.

For instance, to solve the set-theoretic Yang-Baxter equation (Equation \ref{yb_eq}), \textbf{AIDN} treats the problem of finding $f_{R}(\theta)$ as an optimization problem with the following objective function:
\begin{equation}
\label{initial loss}
     \mathcal{L}(f_R):=||(f_{R}(\theta) \times id_{A} )\circ (id_{A} \times f_{R}(\theta)\circ (f_{R}(\theta) \times id_{A} ) - (id_{A} \times f_{R}(\theta)) \circ (f_{R}(\theta) \times id_{A} )\circ  (id_{A} \times f_R(\theta))||_{2}^2.
\end{equation}

 The invertibility of the map $f_R$ can be realized in multiple ways. For example, one may choose to train a neural network that is invertable. This can be done using multiple well-studied methods (e.g., \cite{jacobsen2018revnet,behrmann2019invertible}). Alternatively, one may choose to impose invertibility inside the loss function, define another map $g_R(\alpha)$, and change the loss function in Equation \ref{initial loss} to the following:
\begin{equation*}
\small{
\label{initial loss}
\begin{aligned}
     \mathcal{L}(f_R,g_R)&:=||(f_{R}(\theta) \times id_{A} )\circ (id_{A} \times f_{R}(\theta)\circ (f_{R}(\theta) \times id_{A} )- (id_{A} \times f_{R}(\theta)) \circ (f_{R}(\theta) \times id_{A} )\circ  (id_{A} \times R)||_{2}^2\\ &+ ||f_R(\theta)\circ g_R(\alpha) -id_{A \times A }||_2^2+ ||g_R(\alpha)\circ f_R(\theta)-id_{A \times A }||_2^2.
     \end{aligned}
}  \end{equation*}

We observed that this second method yields more stable solutions in practice. A summary of \textbf{AIDN} algorithm is given in \ref{alg1} below. Further details about the implementation are provided in Appendix \ref{notes}.





\LinesNumbered 
\SetKwProg{Fn}{Function}{}{}
\begin{algorithm}
\Fn{\textbf{AIDN}($\langle S \mid R \rangle$, $k$), $S=\{s_i\}_{i=1}^n$ \textit{is a set of generator and} $R=\{r_i\}_{i=1}^k$  \textit{is a set of relations. Dimension of the representation $k$.}}{
    \ForEach{Generator $s_i$ in $S$}{
        Define the network    $f_i(x;\theta_i) \in \mathcal{N}(\R^k) $\;
        }    
     $\mathcal{L}(f_1,\cdots,f_n ):= \sum_{i=1}^k ||\mathcal{F}(r_i)||_{2}^2 $, where $\mathcal{F}(r_i)$ is the relation $r_i$ written in terms of the networks $\{f_{i}(x;\theta_{i}) \}_{i=1}^n$. \\   
     Minimize $\mathcal{L}(f_1,\cdots,f_n)$ using stochastic gradient descent.\\
    \Return $\{f_i(\theta)\}_{i=1}^n$
    }
    \textbf{End Function}
    \caption{\textbf{AIDN}: Algebraically-Informed Deep Nets }
     \label{alg1}
\end{algorithm}

 Conceptually, Algorithm \ref{alg1} is simple yet it is very general in its applicability to a large set of algebraic objects. Although there is no theoretical guarantee that the neural networks $\{f_{i}(\theta) \}_{i=1}^n$ exist as there is no guarantee that the loss function defined in the \textbf{AIDN} algorithm converges to a global minimum, we consistently found during our experiments that \textbf{AIDN} is capable of achieving good results and finding the desired functions given enough expressive power \cite{cybenko1989approximations,hanin2017approximating,lu2017expressive} for $\{f_{i}(\theta) \}_{i=1}^n$ and enough sample points from the domains of these networks. This observation is consistent with other open research questions in theoretical deep learning \cite{shwartz2017opening} concerning the loss landscape of a deep net. Specifically, multiple research efforts (e.g., \cite{choromanska2015loss,sagun2014explorations}) consistently reported that the parameter landscape of deep networks has a large number of local minima that reliably yield similar levels of performance on multiple experiments. Moreover, the local minima of these landscapes are likely to be close to the global one \cite{choromanska2015loss}.

 Note that the choice of the architecture of the neural network determines the type of the algebraic object representation (e.g. linear, non-linear, etc). In Section \ref{braidgroupsection} and Section \ref{TL}, we will study multiple architectures of neural networks. 
 
 
 In what follows, we explore the applicability of \textbf{AIDN} on different algebraic structures and highlight its properties and performance. All code and data used in this manuscript are publicly available at \textcolor{blue}{\url{https://github.com/mhajij/Algebraically_Informed_Deep_Nets/}}. 





\section{AIDN for Finitely-presented Algebraic Structures}
\label{cases}

Our proposed method is a universal treatment for representing finitely-presented algebraic structures. We demonstrate this generality by applying it to multiple structures that are significant in both geometric topology and algebra including the braid groups (4.1) and the Temperley-Lieb algebras (4.2).


\subsection{Braid Group}
\label{braidgroupsection}

\begin{figure}[h]
  \centering
   {\includegraphics[scale=0.3]{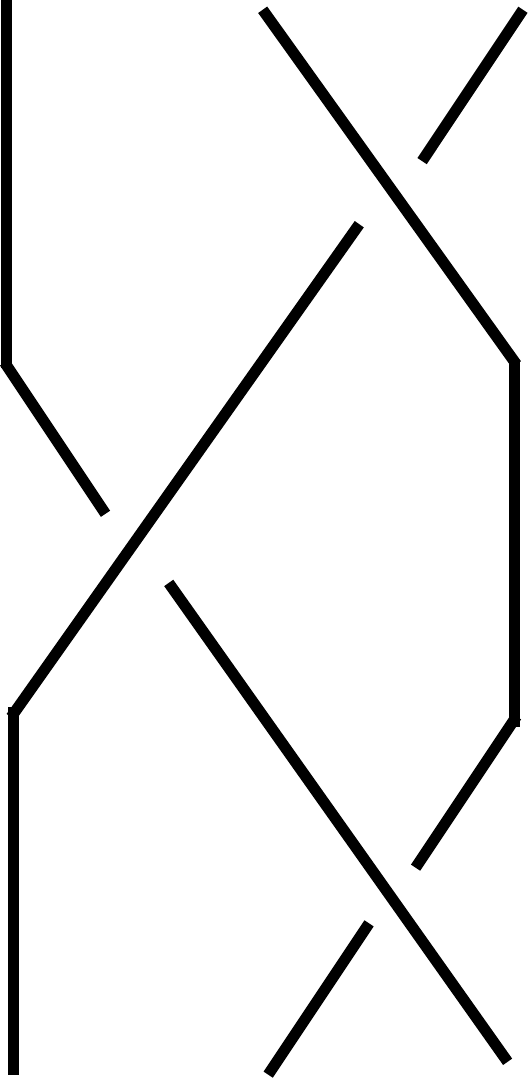}
    \caption{An example of a braid diagram on $3$ strands.}
  \label{braid}}
\end{figure}

Let $D^3$ denotes the cube $[0,1]^3$ in 3-dimensional space, and fix $m$ points on the top face of $D^3$ and $m$ points on the bottom face. A \textit{braid} on $m$ strands is a curve $\beta_m$ embedded in $D^3$ and decomposed into $m$ arcs such that it meets $D^3$ orthogonally in exactly $2m$ points and where no arc intersects any horizontal plane more than once. A braid is usually represented by a planar projection or a \textit{braid diagram}. In the \textit{braid diagram}, we make sure that the over-strand is distinguishable from the under-strand at each crossing by creating a break in the under-strand. Figure~\ref{braid} shows an example of a braid diagram on $3$ strands.

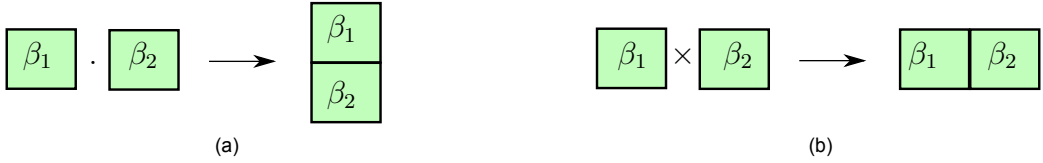
\begin{figure}[h]
  \centering
   {\includegraphics[scale=0.31]{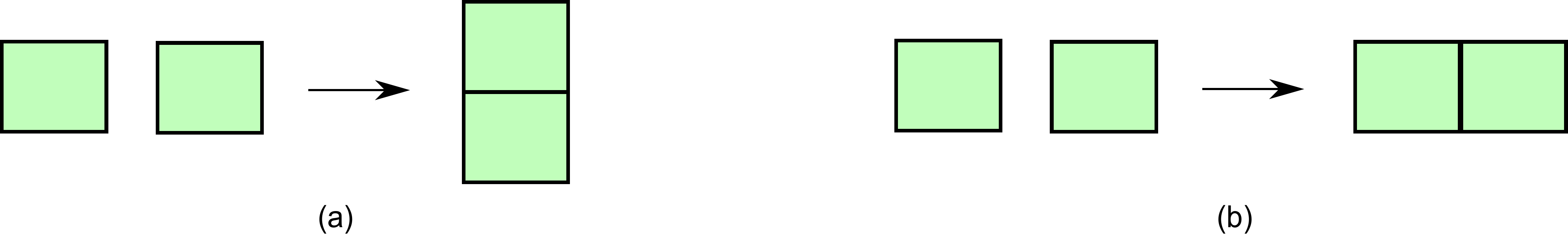}
      \put (-385,35) {\large{$\beta_1$}}
    \put (-345,35) {\large{$\beta_2$}}
     \put (-270,45) {\large{$\beta_1$}}
     \put (-270,20) {\large{$\beta_2$}}
          \put (-360,35) {\large{.}}
      \put (-120,35) {\large{$\beta_2$}}
    \put (-160,35) {\large{$\beta_1$}}
     \put (-20,35) {\large{$\beta_2$}}
     \put (-50,35) {\large{$\beta_1$}}
          \put (-140,35) {\large{$\times$}}          \caption{The product of two braids.}
  \label{product1}}
\end{figure}

The set of all braids $B_m$ has a group structure with multiplication as follows. Given two $m$-strand braids $\beta_1$ and $\beta_2$, the product of these braids ($\beta_1\cdot \beta_2$) is the braid given by the vertical concatenation of $\beta_1$ on top of $\beta_2$ as shown in Figure~\ref{product1}(a).

The group structure of $B_m$ follows directly from this. The braid group $B_m$ on $m$ strands can be described algebraically in terms of generators and relations using Artin's presentation. In this presentation, the group $B_m$ is given by the generators: 
\begin{equation*}
\sigma_1,\ldots,\sigma_{m-1},
\end{equation*}
subject to the relations:
\begin{enumerate}
\item For $|i-j|>1$: $\sigma_i \sigma_j= \sigma _j \sigma _i$. 
\item For all $i<m-1$:  $\sigma_i \sigma_{i+1} \sigma_i = \sigma_{i+1}\sigma_i \sigma_{i+1}$.
\end{enumerate}

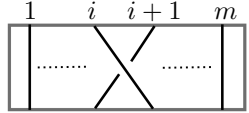
\begin{figure}[h]
  \centering
   {\includegraphics[scale=0.3]{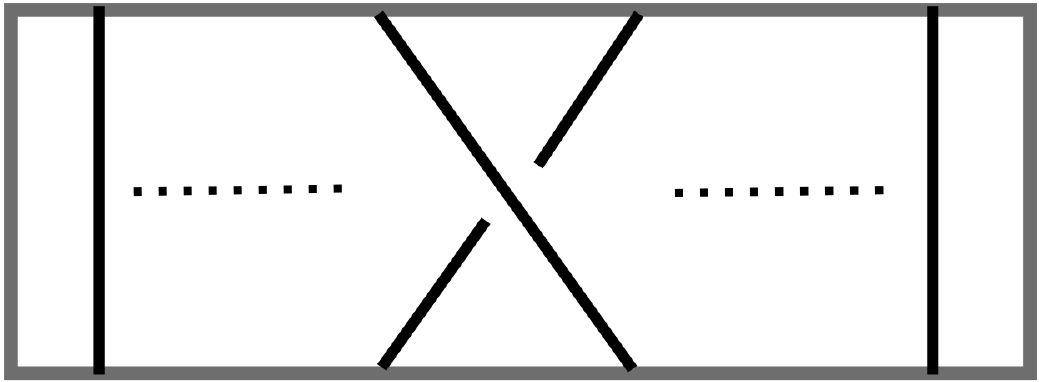}
        \put (-84,35) {\small{$1$}}
    \put (-60,35) {\small{$i$}}
    \put (-45,35) {\small{$i+1$}}
    \put (-12,35) {\small{$m$}}
    \caption{The braid group generator $\sigma_i$.}
  \label{generator}}
\end{figure}

The correspondence between the pictorial definition of the braid group and the algebraic definition is given by sending the generator $\sigma_i$ to the picture illustrated in Figure~\ref{generator}.

\begin{figure}[h]
  \centering
   {\includegraphics[scale=0.078]{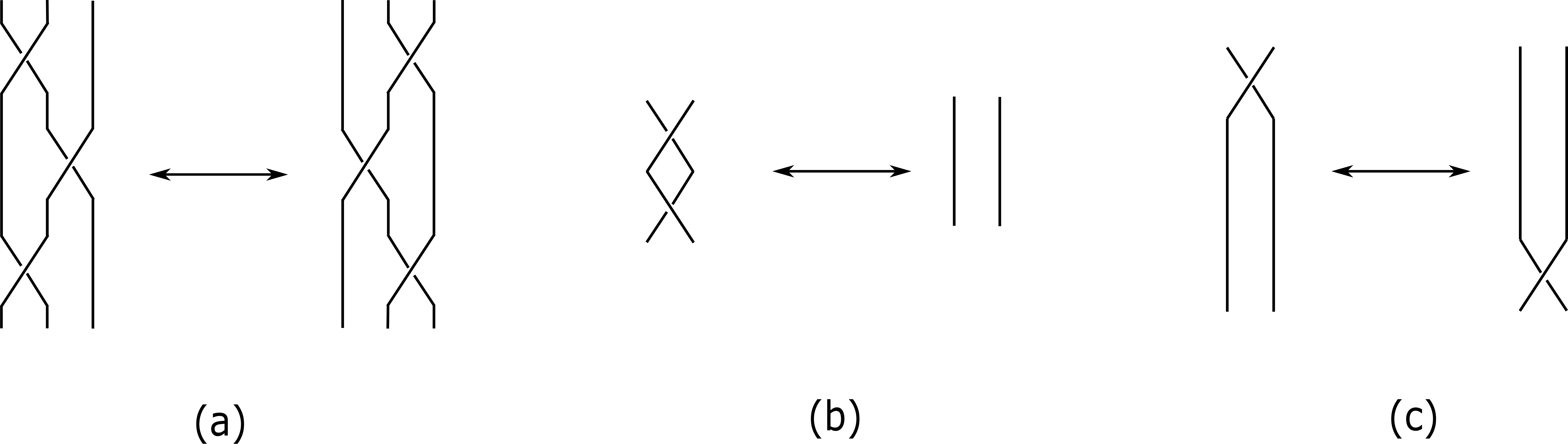}
    \caption{The braid relations.}
  \label{braid relations}}
\end{figure}

Let $\beta_1 \in B_m $ and $\beta _2 \in B_n $.
The product of the braids $\beta_1 $ and $\beta_1 $, denoted by $\beta_1 \times \beta_2 $, is the braid given by horizontal concatenation of  $\beta_1$ to the left of $\beta_2$ as indicated in Figure \ref{product1} (b). If we use $id$ to denote the lone strand and $\sigma$ to denote the crossing appear in the $i$ and $i+1$ position in Figure \ref{generator}, then using the graphical notation given in Figure \ref{product1}(b) we can write: 
\begin{equation}
\label{123}
\sigma_i=
(\times^{i-1} id)  \times \sigma \times (\times^{m-i+1} id), 
\end{equation}
where $\times^k id $ means taking the product of the identity strand $k$ times. This notation will be used throughout the paper. Note that using the graphical notation, the braid relations have intuitive meaning which is illustrated in Figure \ref{braid relations}. In particular, the so-called Reidemeister $3$ corresponds to the relation in Figure 5(a) and Reidemeister $2$ corresponds to the relation in Figure 5(b). Finally, the relation in Figure 5(c) means that the generators can commute as long as they are sufficiently far away from each other.


\subsubsection{Representing the Braid Group Via Neural Networks}
We now have enough background and notations to define the neural networks that can be utilized to represent the braid group. 

Let $B_m$ be the braid group with generators relations as defined earlier. Using \textbf{AIDN}, one might think that we need to train $m-1$ neural networks, since $B_m$ has $m-1$ generators. However, Equation \ref{123} allows us to write every generator in terms of the identity strand, $\sigma$, as well as the product operation given in Figure \ref{product1}.
This observation along with the definition of the braid group implies the following Lemma. 

\begin{lemma}
\label{lamma_braids}
Let $m,n \geq 1 $, $f,g \in \mathcal{N}(\R^n \times \R^n ) $. Define  $f_i \in \mathcal{N}((\R^n)^m )$ by  
\begin{equation}
    f_i:=(\times^{i-1} id_{\R^n} )\times f \times  (\times^{m-i+1} id_{\R^n} ),  
\end{equation}
and define $g_i \in \mathcal{N}((I^n)^m )$ using $g$ similarly. If the functional relations

\begin{enumerate}
\item For all $1 \leq i<m-1$:  $f_{i} f_{i+1} f_{i} = f_{i+1}f_{i} f_{i+1}$
\item For all $1\leq  i < m$:   $f_{i} g_{i} = id_{I^n}=g_{i} f_
{_i} $.
\item For $|i-j|>1$: $f_{_i} f_{_j}= f_{j} f_{i}$. 
\end{enumerate}
are satisfied, then the map $F: B_m \to \mathcal{G}((\R^n)^m ) $ given by 
$   \sigma_i \to f_i$
and $    \sigma^{-1}_i \to g_i $,
and illustrated in Figure \ref{mapp braid}, defines a group homomorphism from $B_m$ to $\mathcal{G}( (\R^n)^m )$.

\begin{figure}[h]
  \centering
   {\includegraphics[scale=0.091]{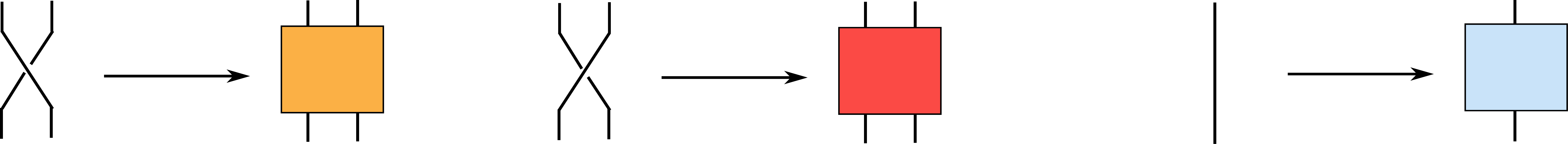}
         \put (-20,44) {\large{$\R^n$}}
      \put (-20,-10) {\large{$\R^n$}}
    \put (-185,44) {\large{$\R^n \times \R^n $}}
    \put (-175,17) {\large{$g$}}    
     \put (-185,-10) {\large{$\R^n \times \R^n $}}
     \put (-20,17) {\large{$Id$}}
    \put (-325,44) {\large{$\R^n \times \R^n $}}
    \put (-315,17) {\large{$f$}}    
     \put (-325,-10) {\large{$\R^n \times \R^n $}}
    \caption{Converting the braid relations into relations between neural networks.}
  \label{mapp braid}}
\end{figure}
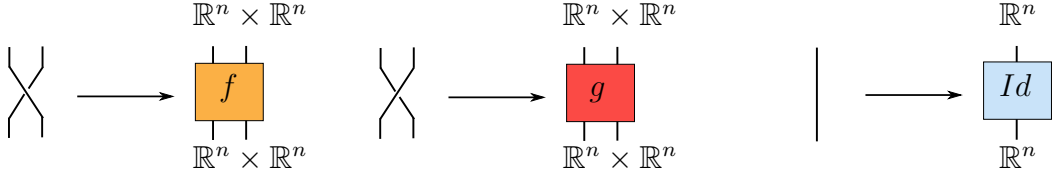
\end{lemma}

Note that in the previous theorem we included the network $g$ in the training process and we train two networks $f$ and $g$ such that they are inverses of each other instead of training a single invertible network $f$. This is because, as we mentioned earlier for the case of the Yang-Baxter equation, the function $f$ needs to be invertible. We found that \textbf{AIDN} performs better if we train two networks that are inverses of each other.

\begin{figure}[h]
  \centering
   {\includegraphics[scale=0.3]{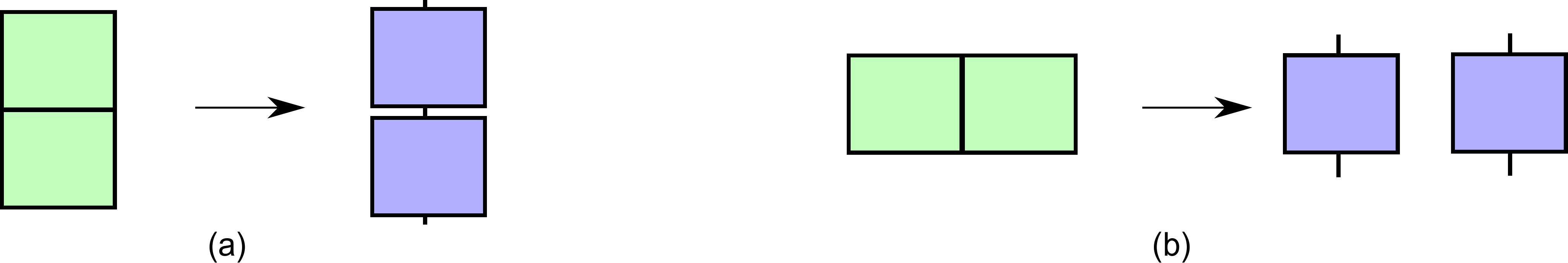}
        \put (-343,42) {\large{$\beta_1$}}
     \put (-343,20) {\large{$\beta_2$}}
     \put (-263,45) {\large{$f_{\beta_1}$}}
     \put (-263,20) {\large{$f_{\beta_2}$}}
      \put (-130,33) {\large{$\beta_2$}}
    \put (-152,33) {\large{$\beta_1$}}
     \put (-22,33) {\large{$f_{\beta_2}$}}
     \put (-59,33) {\large{$f_{\beta_1}$}}
    \caption{(a) Mapping a product of braids to a composition of neural networks. (b) Mapping a product of braids to a product of neural networks.}
  \label{general map}}
\end{figure}
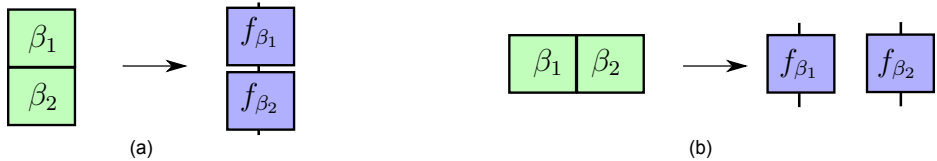

Observe that the map $F$ not only preserves the group structure but also preserves the product operation on braids. Specifically, $F(\beta_1 \times \beta_2  )= F(\beta_1) \times F(\beta_2)$ as illustrated in Figure \ref{general map}.

\begin{figure}[h]
  \centering
   {\includegraphics[scale=0.065]{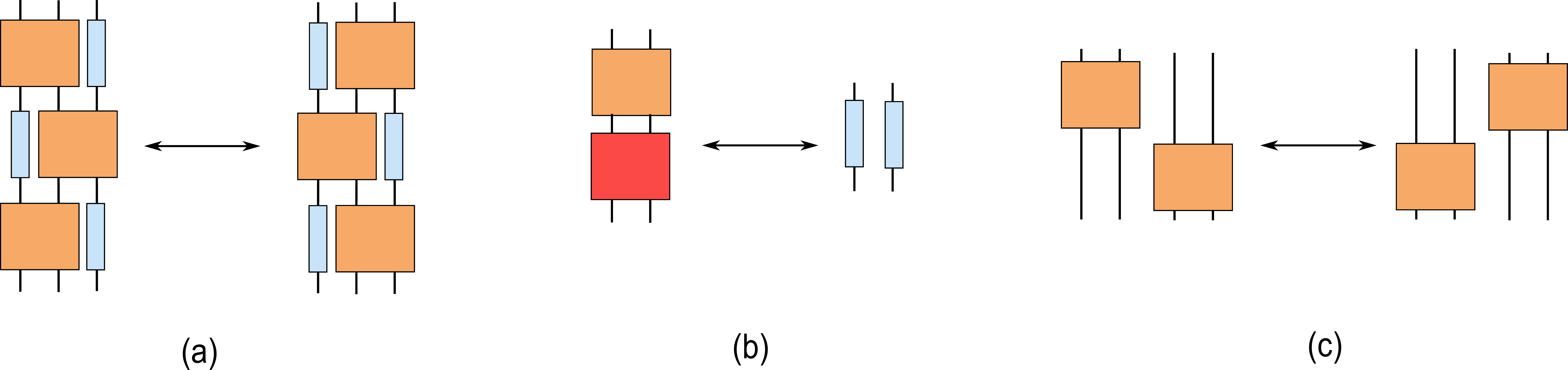}
    \caption{The orange box corresponds to the function $f$ and the red box represents the function $g$. In order for the functions $f$ and $g$ to give us a representation of the braid group, they must satisfy the braid relations given in Figure \ref{braid relations}. (a) This is the functional relation that $f$ must satisfy and corresponds to (a) in Figure \ref{braid relations}. (b) This is the functional relation that $f$ and $g$ must by inverses and corresponds to (b) in Figure \ref{braid relations}. (c) The functional relation that  $f$ must satisfy corresponding to (c) in Figure \ref{braid relations}  }
  \label{rel2}}
\end{figure}

Lemma \ref{lamma_braids} implies that we only need to train two functions $f,g$ that satisfy the relations (a) and (b) given in Figure \ref{rel2} in order to represent any braid group $B_m$. These relations (Figure \ref{rel2}(a--b)) correspond to the braid relations given in Figure \ref{braid relations}(a--b). Observe that the relation in Figure \ref{rel2}(c) is automatically satisfied by our definition of the mapping $F$.



\subsubsection{Performance of \textbf{AIDN} on Braid Group Representations}
\label{results_braids}
Based on Lemma \ref{lamma_braids}, we only need to train two neural networks $f,g \in \mathcal{N}(I^n\times I^n )$ that satisfy the braid relations described in Figure \ref{rel2} to obtain a representation of the braid group. In our experiments, we choose the same architectures for $f$ and $g$. Table \ref{table:braids} reports the results of using the following architecture for networks $f$ and $g$: 
\begin{equation}
\label{choice}
    \R^{n} \to \R^{2n +2} \to \R^{2n +2} \to \R^{100}\to \R^{50}\to\R^{n}.
\end{equation}

Because the choice of the activation function determines the type of representation (linear, affine, or non-linear), we ran three different experiments with three types of networks:
\begin{itemize}
    \item  Linear: In this case, the activation is chosen to be the identity, and we set the bias to be zero for all layers. 
    \item Affine: In this case, the activation is set to the identity with non-zero bias.
    \item Non-linear: In this case, we choose a non-linear activation. Since our choice must guarantee the invariability of the neural network, we performed a hyperparameter search over the possible activation functions and found that the hyperbolic tangent $\tanh$ to give the best results. We also used zero bias for all layers as we found it yields better results. Finally, we used the identity activation for the last layer. 
\end{itemize}

The results of all three cases are reported in Table \ref{table:braids}. 
The results ($L^2$ error) reported in Table \ref{table:braids} are obtained after training the networks for 2 epochs in case of linear and affine representations. In case of non-linear representation, the results ($L^2$ error) are reported after 600 epochs. This, unsurprisingly, indicates the difficulty of training non-linear representations as compared to linear and affine representations.

\begin{table}[t]
\centering
\begin{footnotesize}
  \centering
  \begin{adjustbox}{width=\columnwidth,center}
\begin{tabular}{|c|c|c|c||c|c|c||c|c|c|}
\hline
 \multicolumn{1}{|c|}{}&
 \multicolumn{3}{c||}{$n=2$} & 
    \multicolumn{3}{c||}{$n=4$} & 
 \multicolumn{3}{c|}{$n=6$}\\
\hline
\multicolumn{1}{|c}{braid group relation}&
  \multicolumn{1}{|c|}{Linear} &
  \multicolumn{1}{c|}{Affine} &
    \multicolumn{1}{c||}{Non-Linear} &
  \multicolumn{1}{c|}{Linear} &
  \multicolumn{1}{c|}{Affine} &
  \multicolumn{1}{c||}{Non-Linear} &
 \multicolumn{1}{c|}{Linear} &
 \multicolumn{1}{c|}{Affine} &
 \multicolumn{1}{c|}{Non-Linear}\\
\hline
$f\circ g= id_{I^n}  $ &
$15\times 10^{-6}$& 
$12 \times 10^{-6}$ &
$0.04 $ &
$30\times 10^{-6}$& 
$24 \times 10^{-6}$ &
$0.02$ &
$30\times 10^{-6} $ &
$31\times 10^{-6}$&
$0.04$ \\
\hline
$g\circ f= id_{I^n}$ &
$10\times 10^{-6}$ &
$11\times 10^{-6}$ &
$0.02$ &
$25\times 10^{-6}$ &
$18\times 10^{-6}$ &
$0.02$ &
$32\times 10^{-6} $ &
$30\times 10^{-6}$&
$0.04$  \\
\hline
set-theoretic Yang Baxter &
$75 \times 10^{-7}$ &
$70 \times 10^{-7}$ &
$0.007 $ &
$32\times 10^{-6}$ &
$29\times 10^{-6}$ &
$0.01$ &
$29\times 10^{-6} $ &
$27\times 10^{-6}$&
$0.01$  \\
\hline
\end{tabular}
\end{adjustbox}
\end{footnotesize}
\hspace{5pt}
  \caption{The table describes $L^2$ error of the braid group relations reported after training the networks $f$ and $g$. } 
 \label{table:braids}
\end{table}

\begin{figure}[!h]
  \centering
   {\includegraphics[scale=0.2]{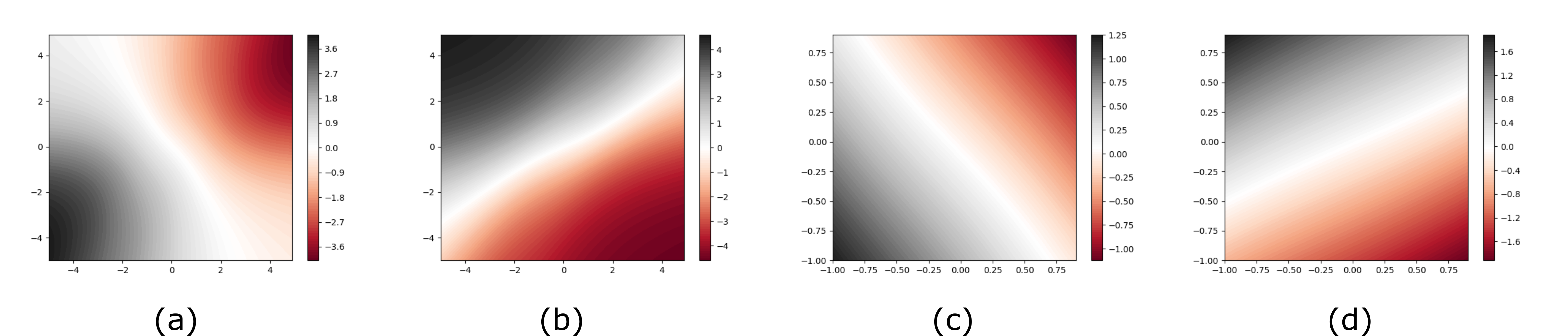}
    \caption{(a) and (b) represent the projections of the trained function $f\in \mathcal{N}(\mathbb{R}^2) $ to the plane. (b) and (c) represent the same function but the region of plotting is constrained to the training region which is $[-1,1]^2$ in this case. Observe that the function is almost linear when constrained on this region. }
  \label{nonlinear}}
\end{figure}

Some solutions obtained on linear braid group representations are given in Table \ref{table:solutions_braidgroup}. We only show the final matrix that corresponds to the function $f$. We note that the solutions given in Table \ref{table:solutions_braidgroup} are solutions for the Yang-Baxter equation. These results are promising and prove the feasibility of using the proposed \textbf{AIDN} for braid group representations. Figure \ref{nonlinear} shows a visualisation of the non-linear solution when $n=2$ by projecting its components to the plane and visualize them as scalar functions.


\begin{table}[h]
\begin{tabular}{|c|c|}
\hline
\textbf{ $n=2$ } & \textbf{$n=4$} \\
\hline
$\left[ \begin{array}{cc}  -0.7115346 &   0.54249334 \\ -0.02051556 &-0.54249316 \end{array}\right]$ &
$\left[ \begin{array}{cccc}  0.27163547&  -0.08972287&   0.30026573&  -0.1708404   \\ 0.2937515 & -0.2939771 & -0.57252926& -0.4431777  \\ 0.36519125&  0.82168174& -0.30026567&  0.17084022  \\ 0.23628233& -0.10712388&  0.572529  &  0.44317824] \end{array}\right]$  \\
\hline
\end{tabular}
  \caption{Some of the solutions obtained by \textbf{AIDN} for linear representations of the braid group. The displayed matrices are obtained from the trained function $f$ after multiplying its weight matrices. } 
 \label{table:solutions_braidgroup}
\end{table}

\subsection{Temperley-Lieb algebra}
\label{TL}
In this second example, we will explore the performance of using \textbf{AIDN} for representing the Temperley-Lieb algebra. Let $\mathcal{R}$ be a ring with a unit, and let $m \geq 1$ be an integer. 

Similar to the braid group, the Temperley-Lieb algebra can be defined via graphical diagrams as follows. Let $D^2$ be the rectangular disk $[0,1]\times [0,1]$. Fix $m$ designated points $\{x_i\}_{i=1}^{m}$ on the top edge of $D^2$, where $x_i=(1,\frac{i}{m+1})$ for $1\leq i \leq m$, and $m$ designated points on the bottom edge $\{x_i\}_{i=m+1}^{2m}$ of $D^2$, where $x_i=(0,\frac{i-m}{m+1})$ for $m+1\leq i \leq 2m$. An $n$-diagram in $D^2$ is a collection of noncrossing embedded curves in $D^2$ drawn on $2m$ designated points such that every designated point in $D^2$ is connected to exactly one other designated point by a single curve. The $m^{th}$ Temperley-Lieb algebra $TL_m$ is the free $\mathcal{R}$-module generated by all $m$-diagrams. This algebra admits a multiplication given by juxtaposition of two diagrams in $[0,1]\times [0,1]$. More precisely, let $D_1$ and $D_2$ be two diagrams in $[0,1] \times[0,1]$ such that $\partial D_j$, where $j=1,2$, consists of the points $\{x_i\}_{i=1}^{2m}$ specified above. Define $D_1.D_2$ to be the diagram in $[0,1]\times[0,1]$ obtained by attaching $D_1$ on the top of $D_2$ and then compress the result to $[0,1]\times[0,1]$. This extends by linearity to a multiplication on $TL_m$. With this multiplication, $TL_m$ is an associative algebra over $\mathcal{R}$.

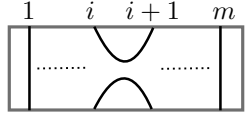
\begin{figure}[h]
  \centering
   {\includegraphics[scale=0.3]{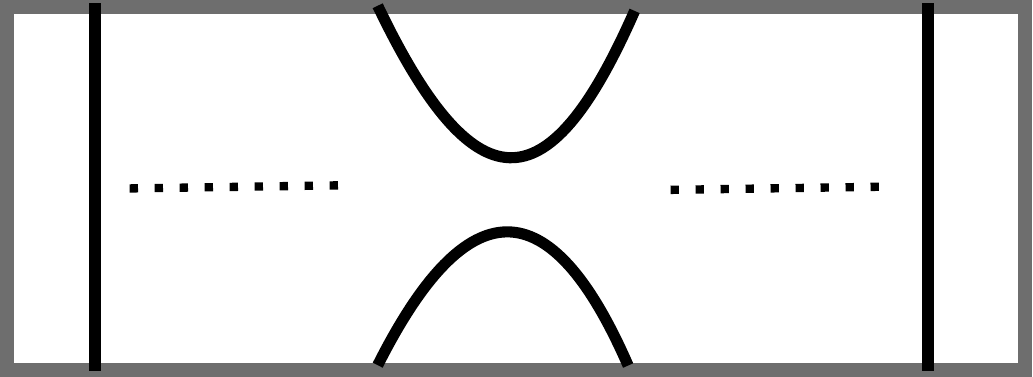}
        \put (-84,35) {\small{$1$}}
    \put (-60,35) {\small{$i$}}
    \put (-45,35) {\small{$i+1$}}
    \put (-12,35) {\small{$m$}}
    \caption{The Temperley-Lieb algebra generator $U_i$.}
  \label{generator2}}
\end{figure}

The $m^{th}$ Temperley-Lieb algebra admits an intuitive presentation in terms of generators and relations as follows. In this presentation, the $m^{th}$ Temperley-Lieb is generated by $m-1$ generators: \[U_1,\cdots, U_{m-1}, \]
where generator $U_i$ is the diagram given in Figure \ref{generator2}.


The generators $U_i$ satisfy the following relations:

\begin{enumerate}

\item For all $1\leq i\leq m-2$:  $U_i U_{i+1} U_i = U_{i}$.
\item For all $2\leq i\leq m-1$:  $U_i U_{i-1} U_i = U_{i}$.
\item For all $1\leq  i < m$:  $U^2_i=\delta U_i $ for all $1\leq i \leq m-1 $
\item For $|i-j|>1$: $U_i U_j= U _j U_i$. 
\end{enumerate}

\begin{figure}[h]
  \centering
   {\includegraphics[scale=0.052]{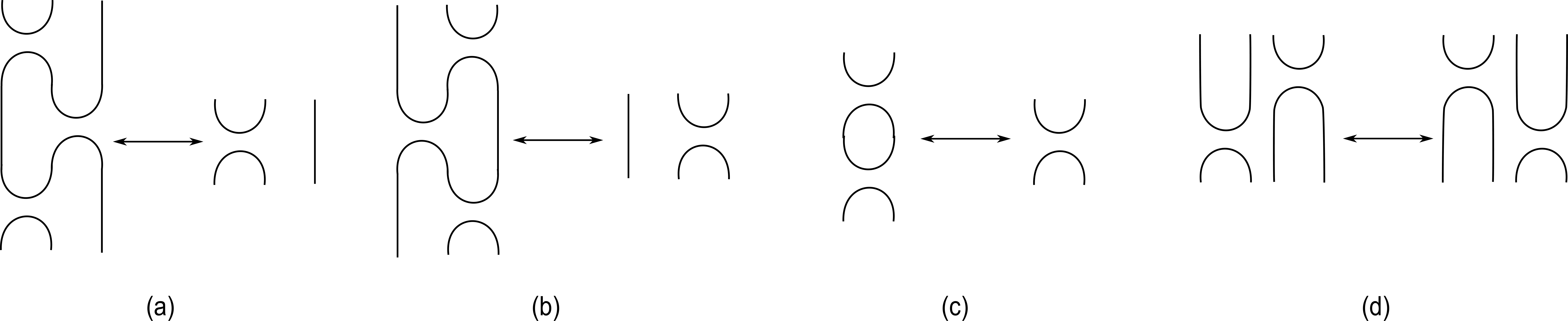}
    \caption{Temperley-Lieb algebra relations.}
  \label{rel}}
\end{figure}

One may observe that with the diagrammatic definition of $U_i$, the relations above have a natural topological interpretation as shown in Figure \ref{rel}.
 
\subsubsection{Representing the Temperley-Lieb  Algebra Via Neural Networks}
\textbf{AIDN} can be utilized to find a representation of $TL_m$ by defining $m-1$ neural networks that satisfy the $TL_m$ relations. Similar to the braid group, we can reduce the number of neural networks that we need to train to one as discussed below. 

Observe that the only difference between the generators $U_i$ is the position of the hook diagram. Specifically, the generator $U_i$ can be built from simple blocks corresponding to the hook and the identity strand. More precisely, define the product of $m$ and $n$ diagrams $D_1 $ and $D_1$, denoted by $D_1 \times D_2 $, to be the $n+m$ diagram given by horizontal concatenation of  $D_1$ to the left of $D_2$. Just as with the braid group, we use $id$ to denote the lone strand and $U$ to denote the hook diagram in the $i$ and $i+1$ position in Figure \ref{generator2} and write $
U_i=
(\times^{i-1} id)  \times U \times (\times^{m-i+1} id). 
$
As before, $\times^k id $ means taking the product of the identity strand $k$ times. With the presentation of the Temperley-Lieb algebra, we immediately arrive at the following lemma: 

\begin{lemma}
\label{lamma2}
Let $m,n \geq 1 $ and $f \in  \mathcal{N}(\R^n \times \R^n) $. Define  $f_i \in \mathcal{N}((\R^n)^m)$, for $1\leq i\leq m-1$, by 
\begin{equation}
    f_i:=(\times^{i-1} id_{\R^n} )\times f \times  (\times^{m-i+1} id_{{\R}^n} ).  
\end{equation}
If the functional relations

\begin{enumerate}
\item For all $1 \leq  i\leq m-2$:  $f_{i} f_{i+1} f_{i} = f_{i}$.
\item For all $2 \leq  i\leq m-1$:  $f_{i} f_{i-1} f_{i} = f_{i}$.
\item For all $1\leq  i < m$:   $f^2_{i}  =\delta f_
{_i} $.
\item For $|i-j|>1$: $f_{_i} f_{_j}= f_{j} f_{i}$. 
\end{enumerate}
are satisfied, then the map $H: TL_m \to \mathcal{N}((\R^n)^m ) $, given by $
    U_i \to f_i,  
$ and illustrated in Figure \ref{mapping},
\begin{figure}[h]
  \centering
   {\includegraphics[scale=0.1]{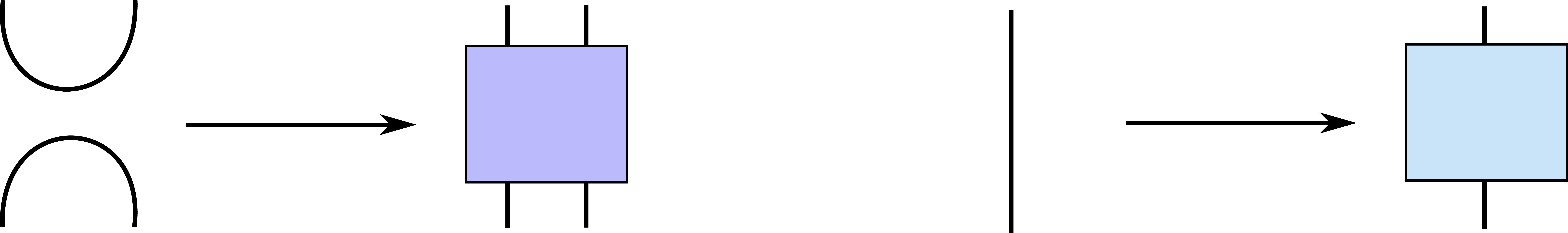}
      \put (-20,44) {\large{$\R^n$}}
      \put (-20,-10) {\large{$\R^n$}}
    \put (-196,44) {\large{$\R^n \times \R^n $}}
    \put (-185,17) {\large{$f_U$}}    
     \put (-196,-10) {\large{$\R^n \times \R^n $}}
     \put (-20,17) {\large{$id$}}
    \caption{We associate the element $U$ in $TL_{2}$ with a trainable neural network $f_{U}: \R^n  \times \R^n \to \R^n  \times \R^n $ and we associate to the single strand in $TL_{1}$ the identity net $id_{\R^{n}}$.}
  \label{mapping}}
\end{figure}
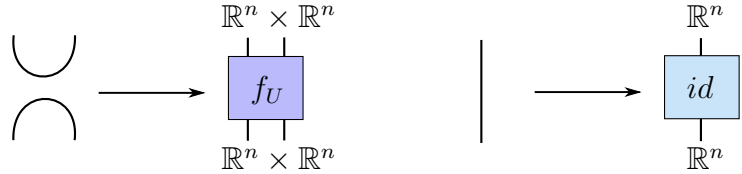
defines an algebra homomorphism from $TL_m$ to $\mathcal{N}( (\R^n)^m )$.

\end{lemma}

\begin{figure}[h]
  \centering
   {\includegraphics[scale=0.06]{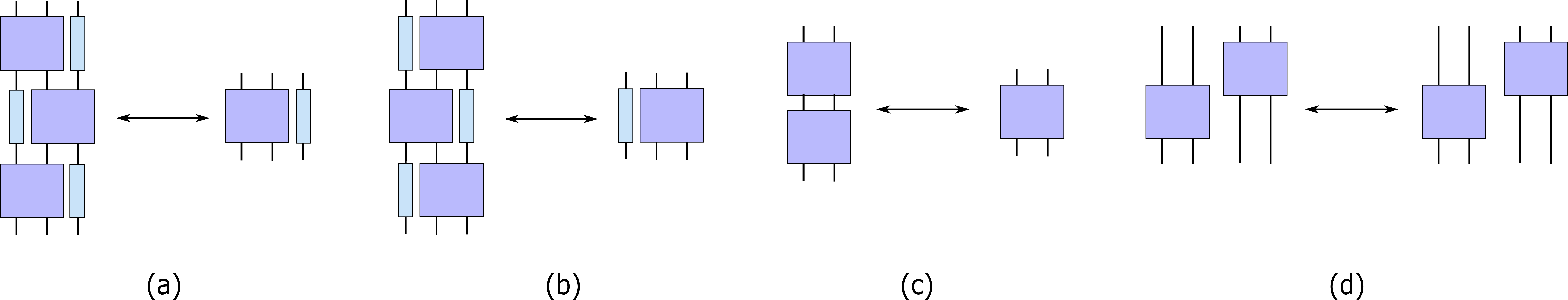}
    \caption{Converting the Temperley-Lieb algebra relations to relations that the neural network $f_U$ must satisfy. The purple box corresponds to the generator $U$. The functional equations in the figure correspond to the relations given in Figure \ref{rel}.}
  \label{rel_2}} 
\end{figure}

Lemma \ref{lamma2} implies that in order to define the mapping $H$, we only need to train a function $f_U$ that satisfies the relations (a), (b) and (c) given in Figure \ref{rel_2}. These correspond to the Temperley-Lieb algebra (a), (b), and (c) given in Figure \ref{rel}. Note that relation \ref{rel_2} (d) is automatically satisfied by our definition of the mapping $H$.

\subsubsection{Performance of \textbf{AIDN} on Temperley-Lieb algebra Representations}
Using Lemma \ref{lamma2}, we are only required to train a single network denoted by $f$. Similar to the experiments of the braid group, we test the \textbf{AIDN} algorithm on the Temperley-Lieb algebra for linear, affine, and non-linear representations. The architecture for the function $U$ is chosen as we did for the braid group generator network in equation \ref{choice}. We also used the same cases defined in Section 4.1.2. The errors ($L^2$) of all the three cases are reported in Table \ref{table:TLalgebra}. 
The linear and affine results reported in Table \ref{table:braids} were obtained by training the network $f$ with only 6 epochs. Training the non-linear representation is harder and required significantly more epochs (600) as compared to the linear and affine representation.


\begin{table}[h]
\centering
\begin{small}
  \centering
  \begin{adjustbox}{width=\columnwidth,center}
\begin{tabular}{|c|c|c|c||c|c|c||c|c|c|}
\hline
 \multicolumn{1}{|c|}{}&
 \multicolumn{3}{c||}{$n=2$} & 
    \multicolumn{3}{c||}{$n=4$} & 
 \multicolumn{3}{c|}{$n=6$}\\
\hline
\multicolumn{1}{|c}{TL relation }&
  \multicolumn{1}{|c|}{Linear} &
  \multicolumn{1}{c|}{Affine} &
    \multicolumn{1}{c||}{Non-linear} &
  \multicolumn{1}{c|}{Linear} &
  \multicolumn{1}{c|}{Affine} &
  \multicolumn{1}{c||}{Non-linear} &
 \multicolumn{1}{c|}{Linear} &
 \multicolumn{1}{c|}{Affine} &
 \multicolumn{1}{c|}{Non-linear}\\
\hline
$U_1 U_2 U_1= U_1  $ &
$81\times 10^{-7} $& 
$64\times 10^{-7}$ &
$0.001$ &
$89\times 10^{-7}$& 
$16 \times 10^{-6}$ &
$0.009$ &
$15\times 10^{-6}$ &
$24\times 10^{-6}$&
$0.005$ \\
\hline
$U_2 U_1 U_2= U_2 $ &
$66\times 10^{-7}$ &
$36\times 10^{-7}$ &
$0.001$ &
$10\times 10^{-6}$ &
$15 \times 10^{-6}$ &
$0.009$ &
$14\times 10^{-6}$ &
$23\times 10^{-6}$&
$0.005$  \\
\hline
$U^2=\delta U$ &
$73\times 10^{-7}$ &
$85\times 10^{-7}$ &
$0.005$ &
$12 \times 10^{-6} $ &
$20 \times 10^{-6}$ &
$0.01$ &
$20\times 10^{-6}$ &
$30\times 10^{-6}$&
$0.01$  \\
\hline
\end{tabular}
\end{adjustbox}
\end{small}
\hspace{5pt}
  \caption{This table describes $L^2$ error of the Temperley-Lieb algebra relations reported after training the network $f$.} 
 \label{table:TLalgebra}
\end{table}

We provide in Table \ref{&} concrete examples of the linear representations of the Temperley-Lieb algebra ($\delta=1$) obtained using \textbf{AIDN}. We also used a plot similar to Figure \ref{nonlinear} and observed that the neural network $f$ with non-linear activations converge to a linear function on the region of training.

\begin{table}[h]
  \begin{adjustbox}{width=\columnwidth,center}
\begin{tabular}{|c|c|}
\hline
\textbf{ $n=2$ } & \textbf{$n=4$} \\
\hline
$\left[ \begin{array}{cc}  0.1648174 & 0.16481737 \\ 0.83518277& 0.83518262 \end{array}\right]$ &
$\left[ \begin{array}{cccc}  0.11085764&  0.1405094 &  0.08143319&  0.1615925   \\  0.15471724&  0.19610043&  0.11365128&  0.22552472 \\ 0.61665326&  0.05473394&  0.91856676& -0.16159254  \\  0.16469169&  0.57503754& -0.1136513 &  0.77447534 \end{array}\right]$  \\
\hline
\end{tabular}
 \end{adjustbox}
  \caption{Some solutions obtained by \textbf{AIDN} for linear representations of the Temperley-Lieb algebra. The displayed matrices are obtained from the trained function $f$ after multiplying its weight matrices. Note that in this case the matrices are their own inverses. } 
 \label{&}
\end{table}

\section{Knot Invariants}
\label{knots}
The study of braid groups is closely related to the study of knot invariants. A \textit{knot} in the $3$-sphere $S^3$ is a smooth one-to-one mapping $f:S^1\to S^3$. A \textit{link} in $S^3$ is a finite collection of knots, called the \textit{components} of the link, that do not intersect with each other. Two links are considered to be equivalent if one can be deformed into the other without any of the knots intersecting itself or any other knots\footnote{This is called ambient isotopy.}. In practice, we usually work with a \textit{link diagram} of a link $L$. A \textit{link diagram} is a projection of $L$ onto $\R^2$ such that this projection has a finite number of non-tangentional intersection points, called crossings. A \textit{link invariant} is a quantity, defined for each link in $S^3$, that takes the same value for equivalent links\footnote{The equivalence relation here is ambient isotopy.}. \textit{Link invariants} play a fundamental role in low-dimensional topology. 

In Section \ref{braidgroupsection}, we showed that braid groups can be defined via neural networks by representing the main building blocks of braids as neural networks and then realizing the braid relations as an optimization problem. Given the close relationship between knots and braids\footnote{Building braid representations is closely related to building link invariants.}, one may wonder if knot invariants can be defined by associating a neural network to each primitive building block of knots and then force the ``knot relations'' on these networks. In this section, we provide an answer to this question. 

\begin{figure}[h]
  \centering
   {\includegraphics[scale=0.3]{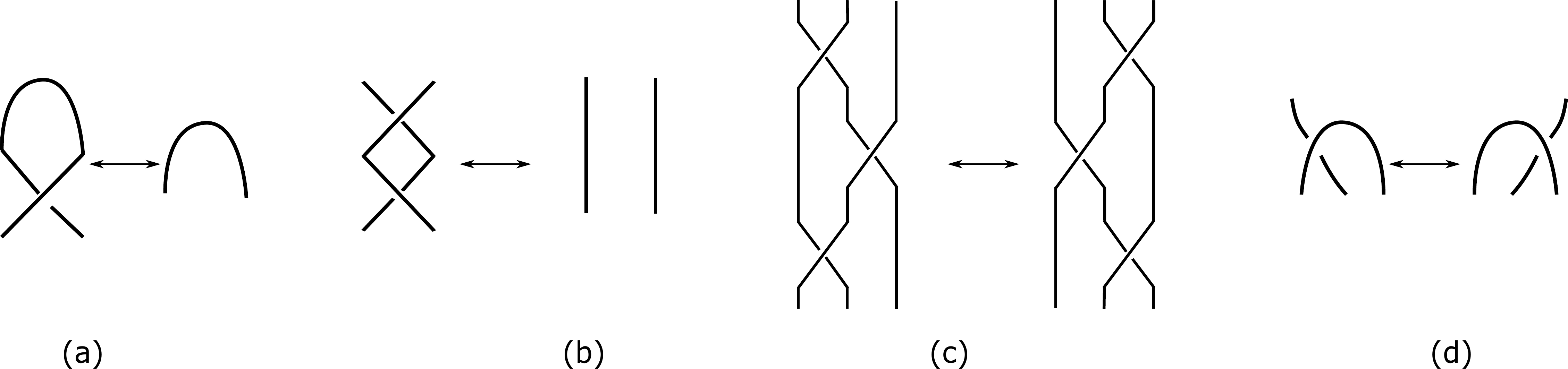}
    \caption{Turaev moves. }
  \label{t moves}}
\end{figure}

In low-dimensional topology, the relation between equivalent knots are called Reidemeister moves \cite{lickorish2012introduction}. Reidemeister moves are similar to braid relations; namely, two link diagrams are equivalent if one diagram can be obtained from the other by a finite sequence of moves of type $\Omega_1$, $\Omega_2$ or $\Omega_3$. These moves are given in Figure \ref{t moves} (a), (b) and (c), respectively. Note that braid relations (a) and (b) are precisely the moves $\Omega_2$ and $\Omega_3$.  

\begin{figure}[t]
  \centering
   {\includegraphics[scale=0.06]{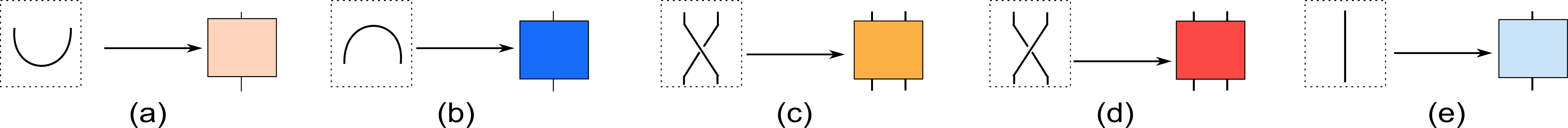}
    \caption{The building blocks of a link are (a) the cup $u:F\to V\otimes V  $, (b) the cap $n :V\times V \to F  $, (c) the simple crossing $R:V\otimes V \to V\otimes V   $, (d) the inverse  $R^{-1}:V\otimes V \to V\otimes V$ of the simple crossing, and (e) identity strand. To build a knot invariant using \textbf{AIDN} we build a neural network for each building block and then we train them to satisfy the relations  in Lemma \ref{inv}.}
  \label{knot mappings}}
\end{figure}

Building knots and links from simple objects requires more primitives than braids. Namely, Turaev \cite{reshetikhin1991invariants} proved that any knot or link can be built with the following primitives: simple crossings, the cup, cap curves, and the identity strand (See Figure \ref{knot mappings}). 
The operations that are needed to build arbitrary knots are similar to those defined on braids in Figure \ref{general map}. Moreover, we require four moves between these building blocks. These moves, which we call the Turaev moves, are presented in Figure \ref{t moves}. We now state the following two lemmas which are due to Reshetikhin and Turaev \cite{reshetikhin1991invariants}.

\begin{lemma}
Every link can be realized as a composition of product of the basic building blocks given in Figure \ref{knot mappings}.
\end{lemma}
The previous Lemma asserts that we can build any link with the basic building blocks. A link diagram built in this way will be called a sliced link diagram. The following lemma can be used to build link invariants from these blocks: 

\begin{lemma}
\label{inv}
\cite{reshetikhin1991invariants,ohtsuki2002quantum}
Let $V$ be a vector space over a field $F$. Let $L$ be a link and $D$ a sliced diagram of $L$.  Let $R$  be  an invertible endomorphism on $ V \otimes V $ and $n:V\otimes V \to F $ a homomorphism that satisfy  (1) $(id_{V} \otimes n)(R\otimes id_{V})=(n\otimes id_{V})(id_{V} \otimes R )$, (2) $n.R=n$, and (3) the Yang-Baxter equation. Then the bracket function $[D]$ defined by maps $R$ and $n$ is an isotopy invariant.  
\end{lemma}

Note that the conditions that $R$ and $n$ must satisfy correspond precisely to the moves given in Figure \ref{t moves}. Using Lemma \ref{inv}, the setup to build a knot invariant using \textbf{AIDN} should now be clear. Specifically, we only need to build four neural networks that correspond to the building blocks given in Figure \ref{knot mappings} and then force the relations specified in Figure \ref{t moves} on these networks. In the quantum invariant literature, the knot invariants obtained using Lemma \ref{inv} are called quantum invariants or Reshetikhin-Turaev invariants \cite{ohtsuki2002quantum}. Hence, \textbf{AIDN}, which can obtain knot invariants, can be considered as a deep learning method to obtain quantum invariants.

In our setting, we must discuss multiple remarks about Lemma \ref{inv}. First, note that unlike the case for the general \textbf{AIDN} algorithm, by Lemma \ref{inv} we must choose the functions $R$, $u$, $n$ to be linear. Second, note that the function $u$ is not strictly required to obtain an invariant using Lemma \ref{inv}. In that context, $u$ can be easily computed using $n$ via the equations $(n\otimes id_{V})(id_{V}\otimes u )=id_{V}=(id_{V}\otimes n ) (u\otimes id_{V}) $ \cite{ohtsuki2002quantum}. However, we found that adding $u$ explicitly to the optimization objective yields better results, so we include these two equations in the optimization process.



\subsection{Performance of AIDN applied to quantum invariants }
Following Lemma \ref{inv}, we use $R$, $R^{-1}$ $u$ and $n$ to denote neural network operators that need to be trained to obtain a RT invariant using \textbf{AIDN}. The equations given in Table \ref{table:RTinvariant} make the final objective function we used to obtain a RT invariant using \textbf{AIDN}.  


\begin{table}[h]
\centering
\begin{small}
  \centering
  \begin{adjustbox}{width=\columnwidth,center}
\begin{tabular}{|c|c|c|}
\hline
 \multicolumn{1}{|c|}{Turaev Move}&
 \multicolumn{1}{c||}{$dim(V)=2$} & 
    \multicolumn{1}{c|}{$dim(V)=4$} \\
\hline
\hline
$n.R= n  $ &
$0.015$& 
$0.044$ \\
\hline
$R\otimes R^{-1}=id_{V\otimes V }   $ &
$0.002$ &
$0.14$  \\
\hline
$R^{-1}\otimes R=id_{V\otimes V }$ &
$0.003$ &
$0.15$  \\
\hline
Yang Baxter&
$0.17$ &
$0.68$  \\
\hline
$(id_{V} \otimes n)(R\otimes id_{V})=(n\otimes id_{V})(id_{V} \otimes R )$ &
$0.07$ &
$0.12$  \\
\hline
$(id_{V} \otimes n)(u\otimes id_{V})=id_{V} $ &
$1.5 \times 10^{-7} $ &
$0.044$  \\
\hline
$(n \otimes id_{V} )(id_{V}\otimes u)=id_{V} $ &
$1.5 \times 10^{-7}$ &
$0.044$  \\
\hline
\end{tabular}
\end{adjustbox}
\end{small}
\hspace{5pt}
  \caption{The table describes $L^2$ error of the RT relations reported after training the networks $R$, $R^{-1}$, $u$, and $n$.} 
 \label{table:RTinvariant}
\end{table}
We make a few comments on Table \ref{table:RTinvariant} in Section \ref{final}.





\section{Conclusion and Future Works}
\label{final}
This work presents \textbf{AIDN}, a novel method to compute representations of algebraic objects using deep learning. We show how the proposed \textbf{AIDN} algorithm can be used to obtain representations of algebraic objects and report the performance of \textbf{AIDN} in finding the solutions for different algebraic structures including groups, associative algebras, and Lie algebras. We also show how to utilize \textbf{AIDN} along with the RT construction to obtain quantum knot invariants. Our experimental results are promising, and open a new paradigm of research where deep learning is utilized to get insights about mathematical problems. We believe this work merely scratch the surface of possibilities of interaction between deep learning and mathematical sciences, and we hope it inspires further research in this direction.

An important remark about the performance of \textbf{AIDN} is the difficulty of training while having many generators and relations. While modern optimization paradigms such as SGD allows one to train a model for high-dimensional data, we found that training multiple networks associated with an algebraic structure with many generators and relations and high-dimensional data to be difficult. This is evident in Table \ref{table:RTinvariant} where the $L^{2}$ error is relatively much higher than the errors obtained while training the braid group and the Temperley-Lieb Algebra networks. This can be potentially addressed using better hyperparameter search and a more suitable optimization scheme.

In the future, we plan to address the following limitations. First, it is not clear whether the neural networks found by \textbf{AIDN} satisfy additional relations that are not explicitly given in the presentation. Intuitively, we would like our searching strategy to learn the exact relations that we provide and be as far as possible from all other possible relations.  In the terminology of representation theory, it is not clear how to guarantee that the algorithm converges to a \textit{faithful} representation of a given dimension, if one does exist.  In our experimentation, we empirically tested the trained neural networks for additional relations that they could potentially satisfy, and found that these networks do not satisfy these potential relations. We did not include these results because this testing approach is not systematic. However, we plan to investigate a systematic test for such a limitation in the future. 

The theory of finite-dimensional (linear) representations of semisimple Lie groups and Lie algebras is very mature, and there are many well-known combinatorial results in this area, including an explicit classification through the famous Highest Weight Theorem \cite{fulton2013representation}. It would be interesting to see whether it could be proven mathematically in this context (or a similar one) that \textbf{AIDN} can be made to converge to a given irreducible representation with a given highest weight. More generally, the theory of (linear) representations of associative algebras, Lie algebras, Jordan algebras, finite groups, Lie groups, etc.\ has been thoroughly investigated over the last century, and many powerful applications have been found to physics, PDEs, harmonic analysis, and several other fields. Hence, it would be useful and interesting to see whether it is possible to guarantee that \textbf{AIDN} converges to a given irreducible representation under certain conditions. 

Finally, since \textbf{AIDN} utilizes SGD to find a representation of a given dimension of some algebraic structure, the obtained solution is not unique. This raises the following question: can we understand the distribution of the local minima solutions obtained by \textbf{AIDN} when applied to a particular algebraic structure? From this perspective, we can consider \textbf{AIDN} as our tool to sample from this unknown distribution. Consequently, \textbf{AIDN} can be potentially utilized in studying general properties of the solution distribution space using other tools available in deep learning such as generative adversarial networks \cite{goodfellow2014generative}.

\section{Acknowledgment}
The authors would like to thank Masahico Saito and Mohamed Elhamdadi for their valuable comments.








\bibliography{main.bib}

\appendix

\section{Appendix}

\subsection{Note on Implementation}
\label{notes}
 
 To highlight the simplicity of the implementation of the \textbf{AIDN} algorithm, we briefly discuss pieces of pseudocode for the study case of the braid group that we discussed in Sections \ref{braid}.
 
 To train a braid group representation using \textbf{AIDN}, we start by creating the generator neural networks $f,g \in \mathcal{N}(\mathbb{R}^n \times \mathbb{R}^n) $, where $n>1$, as explained in Section \ref{results_braids}. To train $f,g$ we create an auxiliary neural network for the relations of the braid group. Specifically, the auxiliary network is trained with the loss function:
 \begin{equation}
     MSE = MSE_{R_2} + MSE_{R_3},       
 \end{equation}
 where 
 \begin{equation}
     MSE_{R_2}= \sum_{i=1,j=1}^{n}||(f\circ g)(x_i,y_j)-(x_i,y_j)||_2^2+||(g\circ f)(x_i,y_j )-(x_i,y_j)||_2^2, 
 \end{equation}
and
\begin{equation}
    MSE_{R_3}=\sum_{i=1,j=1,k=1}^n ||(f \times id)\circ (id \times f) \circ (f \times id) (x_i,y_j,z_k)-(id \times f)\circ (f \times id)\circ (id \times f)(x_i,y_j,z_k) ||_2^2.
\end{equation}
Here, $\{x_i,y_i,z_i\}_{i=1}^n $ denote  points that are sampled uniformly from $\Omega^n \times \Omega^n \times \Omega^n$ where $\Omega \subset \mathbb{R} $, typically the unit interval $[0,1]$. When $n$ is large, a mini-batch setting for stochastic gradient descent is employed for efficiency \cite{goodfellow2016deep,kingma2014adam}. 

\end{document}